\documentclass{article}

\usepackage{authblk}
\usepackage{amsmath}
\usepackage{graphicx}
\usepackage{multirow}

\usepackage[T1]{fontenc}
\usepackage[utf8]{inputenc}
\usepackage{lmodern}

\usepackage{amsthm}
\usepackage{amssymb}
\usepackage{centernot}
\usepackage[hidelinks]{hyperref}
\usepackage{enumitem}

\newtheorem{theorem}{Theorem}
\newtheorem{definition}{Definition}
\newtheorem{example}{Example}

\newcommand{\lin}{{\fontfamily{lmss}\fontshape{n}\selectfont \text{in}}}
\newcommand{\lout}{{\fontfamily{lmss}\fontshape{n}\selectfont \text{out}}}
\newcommand{\lundec}{{\fontfamily{lmss}\fontshape{n}\selectfont \text{undec}}}

\begin{document}

\title{%
An Interleaving Semantics of the Timed Concurrent Language for Argumentation to Model Debates and Dialogue Games%
}

\author[1]{Stefano Bistarelli}
\author[2]{Maria Chiara Meo}
\author[1]{Carlo Taticchi}
\affil[1]{University of Perugia, Italy \texttt{\{stefano.bistarelli,carlo.taticchi\}@unipg.it}}
\affil[2]{University ``G. d'Annunzio'' of Chieti-Pescara, Italy \texttt{mariachiara.meo@unich.it}}

\date{}

\maketitle

\begin{abstract}
Time is a crucial factor in modelling dynamic behaviours of intelligent agents: activities have a determined temporal duration in a real-world environment, and previous actions influence agents' behaviour. In this paper, we propose a language for modelling concurrent interaction between agents that also allows the specification of temporal intervals in which particular actions occur.
Such a language exploits a timed version of Abstract Argumentation Frameworks to realise a shared memory used by the agents to communicate and reason on the acceptability of their beliefs with respect to a given time interval.
An interleaving model on a single processor is used for basic computation steps, with maximum parallelism for time elapsing. Following this approach, only one of the enabled agents is executed at each moment.
To demonstrate the capabilities of language, we also show how it can be used to model interactions such as debates and dialogue games taking place between intelligent agents.
Lastly, we present an implementation of the language that can be accessed via a web interface.
Under consideration in Theory and Practice of Logic Programming (TPLP).

\end{abstract}

\section{Introduction}\label{sec:intro}

Agents in distributed environments can perform operations that affect the behaviour of other components.
Many formalisms have been proposed for modelling concurrent systems to describe the interactions that may take place between intelligent agents. For example, Concurrent Constraint Programming (CC) --~\cite{saraswatConcurrentConstraintProgramming1990} -- relies on a constraint store of shared variables in which agents can read and write in accordance with some properties posed on the variables. The basic operations that agents can execute in the CC framework are a blocking \textit{Ask} and an atomic \textit{Tell}. These operations realise the interaction with the store and also allow one to deal with partial information.

When dealing with concurrent interactions, the notion of time plays a fundamental role: in many ``real-life'' applications, the
activities have a temporal duration that can even be interrupted, and the coordination of such activities has to consider this timeliness property. The interacting actors are
mutually influenced by their actions, meaning that $A$ reacts
accordingly to the timing and quantitative aspects related to $B$'s behaviour, and vice versa. Moreover, the information brought forward by debating agents that interact in a dynamic environment can be affected by time constraints limiting, for instance, the influence of some arguments in the system to a certain time lapse.
Therefore, a mechanism for handling time is required to better model the behaviour of intelligent agents involved in argumentation processes.

Moreover, to simulate human behaviour, intelligent agents operating within distributed systems should communicate with each other through forms of interaction, like debate, that mirror those used by human beings. Debate constitutes, for example, the basis of multi-agent applications for decision making --~\cite{DBLP:conf/ro-man/AzharS16}, planning --~\cite{DBLP:journals/logcom/PardoG18}, opinion polling --~\cite{DBLP:conf/prima/RagoT17}, and negotiation --~\cite{DBLP:conf/ecai/AmgoudPM00}.
We will use the terms dialogue and debate interchangeably, as both can be modelled in the same way and differ only in purpose: the former is typically intended in a collaborative sense, while the latter is usually oppositional.
\cite{walton1995commitment} categorise dialogues into six different forms depending on the information the participants have, their individual goals, and the goals they share:
in an \emph{information-seeking dialogue}, one participant seeks the answer to a question by consulting another participant who presumably knows the answer;
in an \emph{inquiry dialogue}, on the other hand, participants work together to find the answer, which nobody knows, to a question;
in a \emph{persuasion dialogue}, one participant tries to convince the other to accept a proposition;
the participants in a \emph{negotiation dialogue} aim to reach an agreement on the division of a given good or resource;
in a \emph{deliberation dialogue}, participants work together to make a decision about an action to take;
finally, in an \emph{eristic dialogue}, participants aim to emerge victorious by countering the arguments presented by others rather than coming to a truth or conclusion acceptable to all.

A dialogue between two or more parties often unfolds as an ordered series of actions in which the various participants take turns to exchange arguments. With these premises, allowing an agent to have too many choices about which actions to take can be counterproductive and blow up the state space.
It may therefore be convenient to resort to the so-called dialogue games --~\cite{DBLP:books/sp/09/McBurneyP09}, i.e., rule-guided interaction that provides sufficient expressiveness while avoiding state space explosion.
Dialogue games, known since ancient times --~\cite{topics}, find application in different areas, ranging from philosophy to computational linguistics and computer science --~\cite{DBLP:books/sp/09/McBurneyP09}.
In particular, within the field of Artificial Intelligence, they provide valuable support to model, analyse and automatise human-like reasoning.
For example,~\cite{DBLP:journals/aamas/McBurneyEPA03} use dialogue games to model interacting agents that autonomously perform negotiation tasks, while in the work of~\cite{DBLP:journals/ail/PrakkenS98}, dialogue games are used to represent and analyse legal reasoning. 
Therefore, the ability to model debates and dialogue games is a relevant feature of languages aimed at representing the interaction between intelligent agents.

In~\cite{tcla1}, we introduced {\itshape tcla}, a timed extension of the Concurrent Language for Argumentation --~\cite{DBLP:conf/aiia/BistarelliT20,app2,claj} -- which models dynamic interactions between agents using basic actions like $add$, $rmv$, $check$ and $test$, and exploits notions from Argumentation Theory to reason about shared knowledge.
The time extension is based on the hypothesis of {\itshape bounded asynchrony}: the computation takes a bounded period of time and is
measured by a discrete global clock.
Parallel operations are expressed in terms of maximum parallelism.
According to the maximum parallelism policy (applied, for example, in~\cite{DBLP:conf/lics/SaraswatJG94,DBLP:journals/jsc/SaraswatJG96}), at each moment every enabled agent of the system is activated. However, this setting implies the existence of an unbounded number of processors ready to handle the execution of a program.
In~\cite{DBLP:conf/padl/BistarelliMT22}, then, we revised {\itshape tcla} semantics by considering a paradigm where the parallel operator is interpreted in terms of interleaving.
The interleaving approach limits the number of enabled agents executed at a time, mimicking the limited number of available processors as in the real world.
We still assume maximum parallelism for actions depending on time. In other words, time passes for all the parallel processes involved in a computation.
This is accomplished by allowing all the time-only dependent actions, that we identify through $\tau$-transitions, to concurrently run with at most one action manipulating the store, i.e., a $\omega$-transition.
This approach, analogous to that adopted by~\cite{DBLP:journals/tocl/BoerGM04}, differs from that considered by~\cite{DBLP:conf/amast/BusiGZ00} where time does not elapse for timeout constructs.
It is also different from works like~\cite{DBLP:conf/coordination/BistarelliGMS08,DBLP:journals/iandc/BoerGM00} that assume maximum parallelism for all types of action.

This paper extends previous work on the interleaving version of {\itshape tcla} --~\cite{DBLP:conf/padl/BistarelliMT22} -- by providing further details on the functionality and implementation of the language. In detail, the additional contents consist of the following points:
\begin{itemize}
	\item a clarification of the differences between debates and dialogue games;
	\item a formalisation of debates and a translation function which produces a \textit{tcla} program starting from a debate;
	\item a formal methodology for modelling dialogue games as \textit{tcla} programs;
	\item a detailed description of the \textit{tcla} implementation with interleaving, also including a comparison with the maximum parallelism-based version;
	\item examples illustrating the concepts introduced alongside the discussions regarding debates, dialogue games and the implementation.
\end{itemize}

The rest of the paper is organised as follows:
in Section~\ref{sec:background}, we summarise the background notions that will be used to present our proposal;
Section~\ref{sec:tcla} presents the interleaving version of {\itshape tcla}, providing both the syntax and the operational semantics;
in Section~\ref{sec:applications}, we exemplify the use of timed paradigms in the proposed language by modelling a deliberation dialogue and a persuasion dialogue game as {\itshape tcla} processes;
in Section~\ref{sec:implementation}, we describe a working implementation of {\itshape tcla}, and we compare the interleaving version with the maximum parallelism one;
Section~\ref{sec:related} features related works relevant to our study;
Section~\ref{sec:conclusions}, finally, concludes the paper by also indicating possible future research lines.

\section{Background}\label{sec:background}

Argumentation Theory aims to understand and model the natural human fashion of reasoning, allowing one to deal with uncertainty in non-monotonic (defeasible) reasoning.
The building blocks of abstract argumentation are defined by~\cite{dungAcceptabilityArgumentsIts1995}.

\begin{definition}[AFs]\label{def:af}
	Let $U$ be the set of all possible arguments,\footnote{The set $U$ is not present in the original definition of AFs --~\cite{dungAcceptabilityArgumentsIts1995} -- and we introduce it for our convenience to distinguish all possible arguments from the adopted ones.} which we refer to as the ``universe''. An Abstract Argumentation Framework is a pair $\langle Arg,R \rangle$ where $Arg \subseteq U$ is a set of adopted arguments and $R$ is a binary relation on $Arg$ representing attacks among adopted arguments.
\end{definition}

AFs can be represented through directed graphs that we depict using standard conventions. For two arguments $a, b \in Arg$, the notation $(a, b) \in R$ (or, equivalently, $a \rightarrow b$) represents an attack directed from $a$ against $b$.

\begin{definition}[Acceptable Argument]
	Given an AF $F = \langle A,R \rangle$, an argument $a \in A$ is acceptable with respect to $D \subseteq A$ if and only if $\forall b \in A$ such that $(b,a) \in R$, $\exists c \in D$ such that $(c,b) \in R$, and we say that $a$ is \textbf{defended} from $D$.
\end{definition}

We identify the sets of attacking arguments as follows.

\begin{definition}[Attacks]
	Let F = $\langle A,R \rangle$ be an AF, $a\in A$ and $S \subseteq A$. We define the sets
	$a^+ = \{b\in A \mid (a,b)\in R\}$
	and $S^+ = \bigcup_{a \in S} a^+ $.
	Moreover, we define $R_{|a}=\{(a,b)\in R\} $.
\end{definition}

The notion of defence can be used for identifying subsets of ``good'' arguments.
The goal is to establish which are the acceptable arguments according to a certain semantics, namely a selection criterion. Non-accepted arguments are rejected. Different kinds of semantics that reflect desirable qualities for sets of arguments have been studied in works like~\cite{baroniIntroductionArgumentationSemantics2011,dungAcceptabilityArgumentsIts1995}.
We first give the definition for extension-based semantics, namely admissible, complete, stable, semi-stable, preferred, and grounded semantics, denoted with \textit{adm}, \textit{com}, \textit{stb}, \textit{sst}, \textit{prf} and \textit{gde}, respectively, and generically with $\sigma$.

\begin{definition}[Extension-based semantics]\label{def:sem}
	Let $F = \langle Arg,R \rangle$ be an AF. A set $E \subseteq Arg$ is conflict-free in $F$, denoted $E \in S_{cf}(F)$, if and only if there are no $a,b \in E$ such that $(a,b) \in R$. For $E \in S_{cf}(F)$, we have that:
	
	\begin{itemize}[leftmargin=2.4em]
		\item $E \in S_{adm}(F)$ if each $a \in E$ is defended by $E$;
		\item $E \in S_{com}(F)$ if $E \in S_{adm}(F)$ and $\forall a \in Arg$ defended by $E$, $a \in E$;
		\item $E \in S_{stb}(F)$ if $\forall a \in Arg\setminus E $, $\exists b \in E$ such that $(b,a) \in R$;
		\item $E \in S_{sst}(F)$ if $E \in S_{com}(F)$ and $E \cup E^+$ is maximal;%
		\footnote{%
			The set $E \cup E^+$ is also called range of $E$ --~\cite{DBLP:conf/comma/Caminada06}.}
		\item $E \in S_{prf}(F)$ if $E \in S_{adm}(F)$ and $E$ is maximal;
		\item $E \in S_{gde}(F)$ if $E \in S_{com}(F)$ and $E$ is minimal.
	\end{itemize}
	
	Moreover, if $E$ satisfies one of the above properties for a certain semantics, we say that $E$ is an extension of that semantics. In particular, if $E \in S_{adm/com/stb/sst/prf/gde}(F)$, we say that $E$ is an admissible/complete/stable/semi-stable/preferred/grounded extension.
\end{definition}

Besides enumerating the extensions for a certain semantics $\sigma$, one of the most common tasks performed on AFs is to decide whether an argument $a$ is accepted in all extensions of $S_\sigma(F)$ or just in some of them. In the former case, we say that $a$ is \textit{sceptically} accepted with respect to $\sigma$; in the latter, $a$ is instead \textit{credulously} accepted with respect to $\sigma$.

\begin{example}
	In Figure~\ref{fig:af1a} we provide an example of AF where sets of extensions are given for all the mentioned semantics:
	$ S_{cf}(F)$ $=$ $\{\{\}$, $\{a\}$, $\{b\}$, $\{c\}$, $\{d\}$, $\{a,c\}$, $\{a,d\}$, $\{b,d\}\}$,
	$S_{adm}(F)$ $=$ $\{\{\}$, $\{a\}$, $\{c\}$, $\{d\}$, $\{a,c\}$, $\{a,d\}\}$,
	$S_{com}(F)$ $=$ $\{\{a\}$, $\{a,c\}$, $\{a,d\}\}$,
	$S_{prf}(F)$ $=$ $\{\{a,c\}$, $\{a,d\}\}$,
	$S_{stb}(F)$ $=$ $\{\{a,d\}\}$,
	and $S_{gde}(F)$ $=$ $\{\{a\}\}$.
	
	\begin{figure}[htb]
		\centering
		\includegraphics[width=0.55\linewidth]{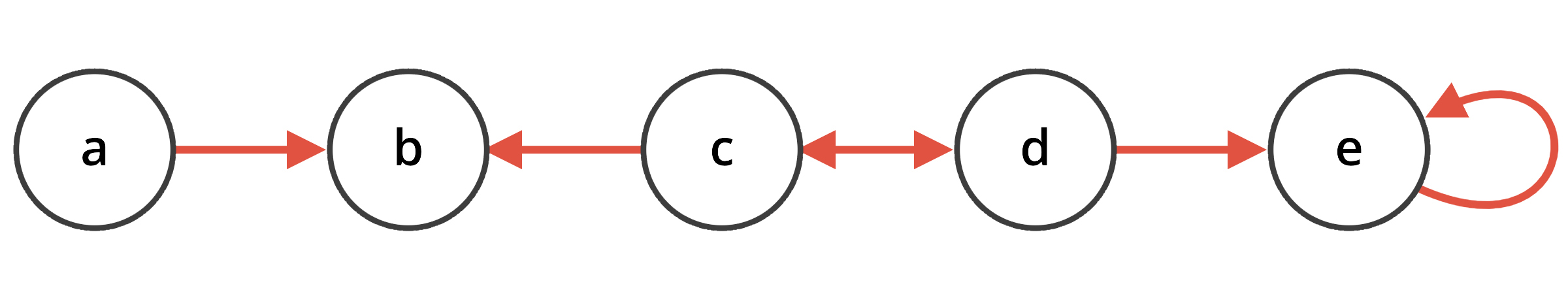}
		\caption{Example of Abstract Argumentation Framework.}
		\label{fig:af1a}
	\end{figure}
	
	The singleton $\{e\}$ is not conflict-free because $e$ attacks itself. The argument $b$ is not contained in any admissible extension because it is never defended from the attack of $a$. The empty set $\{\}$, and the singletons $\{c\}$ and $\{d\}$ are not complete extensions because $a$, which is not attacked by any other argument, has to be contained in all complete extensions. The maximal admissible extensions $\{a,c\}$ and $\{a,d\}$ are preferred, while the minimal complete $\{a\}$ is the (unique) grounded extension. The arguments in $\{a,d\}$ conduct attacks against all the other arguments, namely $b$, $c$ and $e$, thus forming a stable extension.
	To conclude the example, we have that $a$ is sceptically accepted with respect to the complete semantics since it appears in all three subsets of $S_{com}(F)$. On the other hand, arguments $c$ and $d$, each of which is in one complete extension only, are credulously accepted with respect to the complete semantics.
	
\end{example}

The phenomenon for which an argument is accepted in some extension because it is defended by another argument belonging to that extension is known as \textit{reinstatement} - \cite{caminadaIssueReinstatementArgumentation2006}.
Caminada also gives a definition for reinstatement labelling.

\begin{definition}[Reinstatement labelling]\label{def:reinstatement_a}
	Let $\mathit{F}= \langle Arg, R \rangle$ and $\mathbb{L} = \{ \lin, \lout,\lundec\}$. A labelling of $\mathit{F}$ is a total function $L : Arg \rightarrow \mathbb{L}$. We define $in(L) = \{a \in Arg \mid L(a) = \lin \}$, $out(L) = \{a \in Arg \mid L(a) = \lout\}$ and $undec(L) = \{a \in Arg \mid L(a) = \lundec\}$.
	We say that $L$ is a reinstatement labelling if and only if it satisfies the following:
	\begin{itemize}[leftmargin=2.4em]
		\item $\forall a,b \in Arg$, if $a \in in(L)$ and $(b,a) \in R$ then $b \in out(L)$;
		\item $\forall a \in Arg$, if $a \in out(L)$ then $\exists b \in Arg$ such that $b \in in(L)$ and $(b,a) \in R$.
	\end{itemize}
\end{definition}

\begin{figure}[htb]
	\centering
	\includegraphics[width=0.55\linewidth]{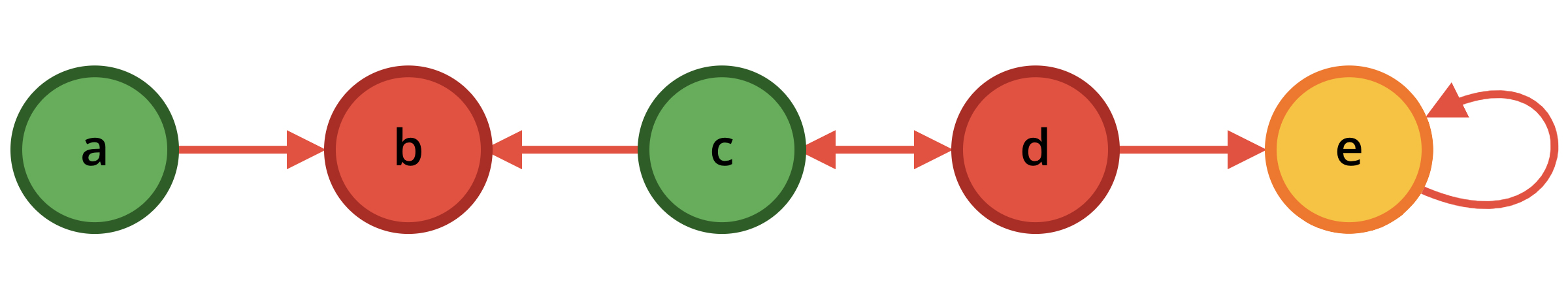}
	\caption{Example of reinstatement labelling.}
	\label{fig:af2a}
\end{figure}

In other words, an argument is labelled \lin\ if all its attackers are labelled \lout, and it is labelled \lout\ if at least one \lin\ node attacks it. In all other cases, the argument is labelled \lundec.
In Figure~\ref{fig:af2a}, we show an example of reinstatement labelling on an AF in which arguments $a$ and $c$ highlighted in green are \lin, red ones ($b$ and $d$) are \lout, and the yellow argument $e$, that attacks itself, is \lundec.

One of the strengths of labelling functions is the possibility of identifying arguments acceptable according to extension-based semantics.
Given a labelling $L$, the set of \lin\ arguments coincides with a complete extension, while other semantics can be obtained through restrictions on the labelling as shown in~\cite{baroniIntroductionArgumentationSemantics2011}.


\section{Syntax and Semantics}\label{sec:tcla}

The syntax of our timed concurrent language for argumentation, {\itshape tcla}, is presented in Table~\ref{tab:CA}. In detail, $P$ denotes a generic process, $C$ a sequence of procedure declarations (or clauses), $A$ a generic agent and $E$ a generic guarded agent, i.e., an agent preceded by a condition that must be satisfied in order to continue the execution.
Moreover $t \in \mathbb{N} \cup \{+\infty\}$.

\begin{table}[htb]
	\begin{align*}
		P &::= let~C~in~A\\
		C &::= p(x)::A \mid C,C\\
		A &::= success \mid failure \mid add(Arg,R) \rightarrow A \mid rmv(Arg,R) \rightarrow A \mid E \mid A \| A\\
		   & ~~~~~~~\mid \exists_x A \mid p(x)\\
		E &::= \mbox{c-}test_{t}(a,l,\sigma) \rightarrow A \mid  \mbox{s-}test_{t}(a,l,\sigma) \rightarrow A \mid check_t(Arg,R) \rightarrow A \mid E+E\\
		  & ~~~~~~~\mid E+_P E \mid E \|_G E
	\end{align*}
	\caption{{\itshape tcla} syntax.}
	\label{tab:CA}
\end{table}

The communication between {\itshape tcla} agents is implemented via shared memory,
similarly to {\itshape cla} --~\cite{DBLP:conf/aiia/BistarelliT20,app2} --
and CC --~\cite{saraswatConcurrentConstraintProgramming1990} --
and opposed to other languages as CSP --~\cite{DBLP:journals/cacm/Hoare78} --
and CCS --~\cite{DBLP:books/sp/Milner80} -- which are based on message passing.
We denote by ${\cal E}$ the class of guarded agents and by 
${\cal E}_0$  the class of guarded agents such that all outermost guards have $t=0$. Note that a Boolean syntactic category could be introduced in replacement of ${\cal E}_0$ to handle guards and allow for finer distinctions.
In a {\itshape tcla} {\em process} $P=let~C~in~A$,
$A$ is the initial agent to be executed in the context of the set of declarations $C$.
Then, a clause defined with $C,C$ corresponds to the concatenation of more procedure declarations.
We will usually write a {\itshape tcla} process $P=let~C~in~A$ as the corresponding agent $A$, omitting $C$ when not required by the context.

The operational model of {\itshape tcla} processes  can be formally described
by a labelled transition system $T= ({\it Conf}, Label, \rightarrow
)$, where we assume that each transition step takes exactly one
time-unit.
Configurations in {\it Conf} are pairs consisting of a
process  and an AF $F = \langle Arg,R \rangle$ 
representing the common knowledge base.
$\mathcal{L} = \{\tau,\omega\}$ is the set of labels that we use to distinguish ``real'' computational steps performed by
processes which have the control (label $\omega$) from the
transitions which model only the passing of time (label $\tau$).
Hence, $\omega$-actions are those performed by processes that modify
the store ($add$, $rmv$), check the store
($check_t$, $\mbox{c-}test_{t}$, $\mbox{s-}test_{t}$), call a procedure, and correspond to exceeding a timeout ($check_0$, $\mbox{c-}test_{0}$, $\mbox{s-}test_{0}$). On the
other hand, $\tau$-actions are those performed by timeout
processes ($check_t$, $\mbox{c-}test_{t}$, $\mbox{s-}test_{t}$) in case they do not have control of the processor.
The transition relation
$\stackrel{\omega}\longrightarrow\subseteq  {\it Conf} \times {\it Conf}$ is the least
relation satisfying the rules in Tables~\ref{tab:CAop1} and \ref{tab:CAop2}, and it
characterises the (temporal) evolution of the system. So, 
$\langle A, F \rangle 
\stackrel{\omega}\longrightarrow
\langle A', F' \rangle$ means that, if
at time $t$ we have the process $A$ and the AF $F$, then at
time $t+1$ we have the process $A'$ and the AF $F'$.

In Tables~\ref{tab:CAop1} and \ref{tab:CAop2}, we give the definitions for the transition rules.
The agents $success$ and $failure$ represent a successful and a failed termination, respectively, so they may not make any further transition.
Action prefixing is denoted by $\rightarrow$, non-determinism is
introduced via the guarded choice construct $E+E$, if-then-else statements can be realised through $+_P$, parallel and guarded parallel
compositions are denoted by $\parallel$ and $\parallel_G$, and a notion  of locality is introduced by the agent $\exists_x A$, which behaves like $A$
with argument $x$ considered local to $A$, thus hiding the information on
$x$ provided by the external environment.
Moreover, we have the $\mbox{c-}test_{t}(a,l,\sigma) \rightarrow A$, $ \mbox{s-}test_{t}(a,l,\sigma) \rightarrow A$ and  $check_t(Arg,R) \rightarrow A$ constructs, which are explicit timing primitives introduced in order to allow for the specification of timeouts.
In Tables~\ref{tab:CAop1} and~\ref{tab:CAop2} we have omitted the symmetric rules for the choice operator $+$ and for
the two parallel composition operators $\|$ and $\|_G$.
Indeed, $+$ is commutative, so $E_1 + E_2$ produces the same result as (that is, is congruent to) $E_2 + E_1$. The same is true for $\|$ and $\|_G$. Note that $+$, $\|$ and $\|_G$ are also associative. 
Moreover, $success$ and $failure$ are the identity and the absorbing elements under the parallel composition $\|$, respectively; that is, for each agent $A$, we have that $ A \| success$ and $ A \| failure$ are the agents $A$ and $failure$, respectively.

\begin{table}[htbp]
	\centering
	\small
	\def\arraystretch{3}
	\begin{tabular}{c r}
		$\genfrac{}{}{0pt}{0}{
			\langle add(Arg',R') \rightarrow A, \langle Arg, R \rangle \rangle
			\stackrel{\omega}\longrightarrow 
			\langle A, \langle Arg \cup Arg', R \cup R'' \rangle \rangle
		}
		{
			\text{where } R'' = \{(a,b) \in R' \mid a,b \in Arg \cup Arg'\}
		}$
		& Add\\
		$\genfrac{}{}{0pt}{0}{
			\langle rmv(Arg',R') \rightarrow A, \langle Arg, R \rangle \rangle
			\stackrel{\omega}\longrightarrow
			\langle A, \langle Arg \setminus Arg', R \setminus \{R' \cup R''\} \rangle \rangle
		}
		{
			\text{where } R'' = \{(a,b) \in R \mid a \in Arg' \vee b \in Arg'\}
		}$
		& Rmv\\
		$\genfrac{}{}{0.5pt}{0}{
			Arg' \subseteq Arg \wedge R' \subseteq R
			\ \ \ \ t>0
		}
		{
			\langle check_t(Arg',R') \rightarrow A, \langle Arg,R \rangle \rangle 
			\stackrel{\omega}	\longrightarrow
			\langle A, \langle Arg,R \rangle \rangle
		}$
		& Chk (1)\\
		$\genfrac{}{}{0.5pt}{0}{
			Arg' \not\subseteq Arg \vee R' \not\subseteq R
			\ \ \ \ t>0
		}
		{
			\langle check_t(Arg',R') \rightarrow A, \langle Arg,R \rangle \rangle 
			\stackrel{\omega}	\longrightarrow
			\langle check_{t-1}(Arg',R') \rightarrow A, \langle Arg,R \rangle \rangle
		}$
		& Chk (2)\\
		$\langle check_t(Arg',R') \rightarrow A,F \rangle  	\stackrel{\tau}\longrightarrow
		\langle check_{t-1}(Arg',R') \rightarrow A,F \rangle\ \ \ \ \ t>0$
		&  Chk (3)\\
		$\langle check_0(Arg',R') \rightarrow A,F \rangle 
		\stackrel{\omega}\longrightarrow
		\langle failure, F \rangle$
		&  Chk (4)\\
		$\genfrac{}{}{0.5pt}{0}{
			\exists L \in S_\sigma(F) \mid l\in L(a)
			\ \ \ \ t>0
		}
		{
			\langle \mbox{c-}test_{t}(a,l,\sigma) \rightarrow A,F \rangle 
			\stackrel{\omega}	\longrightarrow
			\langle A,F \rangle
		}$
		& CrT (1)\\
		$\genfrac{}{}{0.5pt}{0}{
			\forall L \in S_\sigma(F) \mid l\not\in L(a)
			\ \ \ \ t>0
		}
		{
			\langle \mbox{c-}test_{t}(a,l,\sigma) \rightarrow A,F \rangle 
			\stackrel{\omega}	\longrightarrow
			\langle \mbox{c-}test_{t-1}(a,l,\sigma) \rightarrow A,F \rangle
		}$
		& CrT (2)\\
		$\langle \mbox{c-}test_{t}(a,l,\sigma) \rightarrow A,F \rangle   	\stackrel{\tau}\longrightarrow
		\langle \mbox{c-}test_{t-1}(a,l,\sigma) \rightarrow A,F \rangle \ \ \ \ t>0$
		&  CrT (3)\\
		$\langle \mbox{c-}test_{0}(a,l,\sigma) \rightarrow A,F \rangle 
		\stackrel{\omega}\longrightarrow
		\langle failure, F \rangle$
		&  CrT (4)\\
		$\genfrac{}{}{0.5pt}{0}{
			\forall L \in S_\sigma(F) . l\in L(a)
			\ \ \ \ t>0
		}
		{
			\langle \mbox{s-}test_{t}(a,l,\sigma) \rightarrow A,F \rangle 
			\stackrel{\omega}	\longrightarrow
			\langle A,F \rangle
		}$
		& ScT (1)\\
		$\genfrac{}{}{0.5pt}{0}{
			\exists L \in S_\sigma(F) . l\not\in L(a)
			\ \ \ \ t>0
		}
		{
			\langle \mbox{s-}test_{t}(a,l,\sigma) \rightarrow A,F \rangle 
			\stackrel{\omega}	\longrightarrow
			\langle \mbox{s-}test_{t-1}(a,l,\sigma) \rightarrow A,F \rangle
		}$
		& ScT (2)\\
		$\langle \mbox{s-}test_{t}(a,l,\sigma) \rightarrow A,F \rangle   	\stackrel{\tau}\longrightarrow
		\langle \mbox{s-}test_{t-1}(a,l,\sigma) \rightarrow A,F \rangle \ \ \ \ t>0$
		& ScT (3)\\
		$\langle \mbox{s-}test_{0}(a,l,\sigma) \rightarrow A,F \rangle 
		\stackrel{\omega}\longrightarrow
		\langle failure, F \rangle$
		& ScT (4)\\
		$\genfrac{}{}{0.5pt}{0}{
			\langle	E_1, F  \rangle \stackrel{\omega}\longrightarrow \langle A_1, F \rangle,\ 
			\langle E_2, F \rangle 
			\stackrel{\tau}\longrightarrow\langle A_2, F\rangle
			\ \ E_1,E_2 \not \in {\cal E}_0,
			\ \ A_1 \not \in {\cal E}
		}
		{
			\langle E_1 \|_G E_2, F \rangle
			\stackrel{\omega}\longrightarrow
			\langle A_1 \| A_2, F \rangle
		}$
		& GPa (1)\\
		$\genfrac{}{}{0.5pt}{0}{
			\langle E_1, F \rangle 
			\stackrel{\xi}\longrightarrow\langle E_1', F\rangle,\ 
			\langle E_2, F \rangle 
			\stackrel{\tau}\longrightarrow\langle E_2', F\rangle,\ 
			\ \ E_1, E_2 \not \in {\cal E}_0,
			\ \ E'_1, E'_2 \in {\cal E}
		}
		{
			\langle E_1  \|_G E_2, F \rangle
			\stackrel{\xi}\longrightarrow
			\langle E'_1  \|_G E'_2, F \rangle
		}
		\ \ \ \  \xi \in \{\tau, \omega\}$
		& GPa (2)\\
		$\genfrac{}{}{0.5pt}{0}{
			E_1 \in {\cal E}_0,
			\langle	E_2, F  \rangle \stackrel{\xi}\longrightarrow\langle A_2, F \rangle
		}
		{
			\langle E_1  \|_G E_2, F \rangle
			\stackrel{\xi}\longrightarrow
			\langle A_2 , F \rangle
		}
		\ \ \ \ \xi \in \{\tau, \omega\}$
		& GPa (3)%
		\vspace{1em}
	\end{tabular}
	\caption{{\itshape tcla} operational semantics (part I).}
	\label{tab:CAop1}
\end{table}

In the following we give an operational semantics of {\itshape tcla}, where  the parallel operator is modelled in terms of {\itshape interleaving}.
While in the {\itshape maximum parallelism} paradigm, at each moment, every enabled agent of the system is activated, in the interleaving paradigm, agents may have to wait for the processor to be ``free''.
Clearly, since we have dynamic process creation, a maximum parallelism approach has the disadvantage that, in general, it implies the existence of an unbound number of processes.
On the other hand a naive interleaving semantic could be problematic from the time viewpoint, as in principle the time does not pass for enabled agent which are not scheduled.

\begin{table}[htb]
	\centering
	\small
	\def\arraystretch{3.1}
	\begin{tabular}{c r}
		$\genfrac{}{}{0.5pt}{0}{
			\langle	E_1, F  \rangle \stackrel{\omega}\longrightarrow \langle A, F \rangle,\ 
			\ \ E_1 \not \in {\cal E}_0,
			\ \ A_1 \not \in {\cal E}
		}
		{
			\langle E_1 +_P E_2, F \rangle 
			\stackrel{\omega}\longrightarrow 
			\langle A_1, F \rangle
		}$
		& ITE (1)\\
		$\genfrac{}{}{0.5pt}{0}{
			\langle E_1, F \rangle 
			\stackrel{\xi}\longrightarrow\langle E_1', F\rangle,\ 
			\ \ E_1 \not \in {\cal E}_0,
			\ \ E'_1 \in {\cal E}
		}
		{
			\langle E_1 +_P E_2, F \rangle
			\stackrel{\xi}\longrightarrow
			\langle E'_1 +_P E_2, F \rangle
		}$
		& ITE (2)\\
		$\genfrac{}{}{0.5pt}{0}{
			E_1 \in {\cal E}_0,
			\langle	E_2, F  \rangle \stackrel{\xi}\longrightarrow \langle A_2, F \rangle\ 
		}
		{
			\langle E_1 +_P E_2, F \rangle 
			\stackrel{\xi}\longrightarrow
			\langle A_2, F \rangle
		}
		\ \ \	\ \xi \in \{\tau, \omega\}$
		& ITE (3)\\
		$\genfrac{}{}{0.5pt}{0}{
			\langle A_1, F \rangle 
			\stackrel{\xi}	\longrightarrow
			\langle A_1', F' \rangle, \ \langle A_2, F \rangle 
			\stackrel{\tau}\longrightarrow
			\langle A'_2, F \rangle 
		}
		{
			\langle A_1 \| A_2, F \rangle
			\stackrel{\xi}\longrightarrow
			\langle A_1' \| A_2', F' \rangle
		}
		\ \ \ \ \xi\in\{\tau,\omega\}$
		& Par (1)\\
		$\genfrac{}{}{0.5pt}{0}{
			\langle A_1, F \rangle 
			\stackrel{\xi}	\longrightarrow
			\langle A_1', F' \rangle,\  \langle A_2, F \rangle 	\not\stackrel{\tau}\longrightarrow
		}
		{
			\langle A_1 \| A_2, F \rangle
			\stackrel{\xi}\longrightarrow
			\langle A_1' \| A_2, F' \rangle  
		}
		\ \ \ \ \xi\in\{\tau,\omega\}$
		& Par (2)\\
		$\genfrac{}{}{0.5pt}{0}{
			\langle E_1, F \rangle 
			\stackrel{\omega}	\longrightarrow
			\langle A_1, F \rangle,\ 
			E_1, E_2 \not \in {\cal E}_0
			\ \ \ \  A_1 \not \in {\cal E}
		}
		{
			\langle E_1 + E_2, F \rangle
			\stackrel{\omega}\longrightarrow
			\langle A_1, F \rangle
		}$
		& NDt (1)\\
		$\genfrac{}{}{0.5pt}{0}{
			E_1 \in {\cal E}_0,
			\langle	E_2, F  \rangle \stackrel{\xi}\longrightarrow\langle A_2, F \rangle
		}
		{
			\langle E_1  + E_2, F \rangle
			\stackrel{\xi}\longrightarrow
			\langle A_2 , F \rangle
		}
		\ \ \ \ \xi \in \{\tau, \omega\}$
		& NDt (2)\\
		$\genfrac{}{}{0.5pt}{0}{
			\langle E_1, F \rangle 
			\stackrel{\xi}	\longrightarrow
			\langle E'_1, F \rangle,\ 
			\langle E_2, F \rangle 
			\stackrel{\tau}	\longrightarrow \langle E'_2, F \rangle
			\ \ \ \ 
			E_1, E_2 \not \in {\cal E}_0,
			\ 	E'_1, E'_2\in {\cal E}
		}
		{
			\langle E_1 + E_2, F \rangle
			\stackrel{\xi}\longrightarrow
			\langle E'_1 + E'_2, F \rangle
		}
		\ \ \ \ \xi \in \{\tau, \omega\}$
		& NDt (3)\\
		$\genfrac{}{}{0.5pt}{0}{
			\langle A[y/x], F \rangle 
			\stackrel{\xi}\longrightarrow
			\langle A', F' \rangle 
		}
		{
			\langle \exists_x A, F \rangle 
			\stackrel{\xi}	\longrightarrow
			\langle A', F' \rangle
		}$
		\ \ \ \ with $y$ fresh 	$\ \ \ \ \xi\in\{\tau,\omega\}$
		& HVa\\
		$ \langle p(y), F \rangle 
		\stackrel{\omega}	\longrightarrow
		\langle A[y/x], F \rangle
		\ \ \ \ $ with $p(x) :: A$ and $x \in \{a,l,\sigma,t\}$
		& PrC
	\end{tabular}
	\caption{{\itshape tcla} operational semantics (part II).}
	\label{tab:CAop2}
\end{table}

For the operational semantics  of {\itshape tcla} we follow a solution
analogous to that adopted in~\cite{DBLP:conf/coordination/BistarelliGMS08}: we
assume that the parallel operator is interpreted in terms of
interleaving, as usual, however we must assume maximum parallelism
for actions depending on time. In other words, time passes for all
the parallel processes involved in a computation. Practically, we use $\tau$-actions to make the time pass for agents who do not require the processor.

Suppose we have an agent  $A$ whose knowledge base is represented by an AF $F = \langle Arg,R \rangle$.
An $add(Arg',R')$ action performed by the agent results in the addition of a set of arguments $Arg' \subseteq U$, where $U$ is the universe, and a set of relations $R'$ to the AF $F$.
When performing an Add, (possibly) new arguments are taken from $U \setminus Arg$.
We want to make clear that the tuple $(Arg',R')$ is not an AF, indeed it is possible to have $Arg' = \emptyset$ and $R' \neq \emptyset$, which allows to perform an addition of only attacks to the considered AF. It is as well possible to add only arguments to $F$, or both arguments and attacks.
Intuitively, $rmv(Arg,R)$ allows to specify arguments and/or attacks to remove from the knowledge base.
Removing an argument from an AF requires to also remove the attacks involving that argument and trying to remove an argument or an attack which does not exist in $F$ will have no consequences.

The operator $check_t(Arg',R')$ realises a timed construct used to verify whether the specified arguments and attacks are contained in the knowledge base, at the time of execution or some subsequent instant before the timeout, without introducing any further change. If $t>0$ and the check is positive, the operation succeeds and the agent $check_t(Arg',R') \rightarrow A$ can
perform a $\omega$-action in the agent $A$ (Rule  Chk (1)).
If $t>0$
and the check is not satisfied, then the control is repeated at the
next time instant and the value of the counter $t$ is decreased; note that in this case we use the label
$\omega$, since a check on the store has been performed  (Rule  Chk (2)). As shown
by axiom  Chk (3), the counter can be  decreased also by
performing a $\tau$-action: intuitively, this rule is used to model
the situation in which, even though the evaluation of the timeout
started already, another parallel process has the control. In
this case, analogously to the approach in~\cite{DBLP:journals/tocl/BoerGM04} and differently from the approach in~\cite{DBLP:conf/amast/BusiGZ00}, time
continues to elapse, via $\tau$-actions, also for the timeout
process. Axiom
Chk (4) shows that, if the timeout is exceeded, i.e., the
counter $t$ has reached the value of $0$, then the process $check_t(Arg',R') \rightarrow A$ fails.

The rules CrT (1)-(4) and ScT (1)-(4) in Table~\ref{tab:CAop1} perform a credulous and a sceptical test, respectively, and are similar to rules Chk (1)-(4) described before.
Observe that we have two distinct test operations, both requiring the specification of an argument $a \in A$, a label $l \in \{\lin, \lout, \lundec\}$ and a semantics $\sigma \in \{adm, com, stb, prf, gde\}$. The credulous $\mbox{c-}test_t(a,l,\sigma)$,
with $t>0$, succeeds if there exists at least one extension of $S_\sigma(F)$ whose corresponding labelling $L$ is such that $L(a) = l$.
Similarly, the sceptical $\mbox{s-}test_t(a,l,\sigma)$, with $t>0$, succeeds if $a$ is labelled $l$ in all possible labellings $L \in S_\sigma(F)$.

The operator $+_P$ (rules ITE (1)-(3)) is left-associative and realises an if-then-else construct: if we have $E_1 +_P E_2$ and $E_1$ is successful, than $E_1$ will be always chosen over $E_2$, even if $E_2$ is also successful, so in order for $E_2$ to be selected, it has to be the only one that succeeds. The guarded parallelism $\|_G$ (rules GPa (1)-(3)) is designed to allow all the operations for which the guard in the inner expression is satisfied. In detail, $E_1 \|_G E_2$ is successful when either $E_1$, $E_2$ or both are successful and all the operations that can be executed are executed. This behaviour is different both from classical parallelism (for which all the agents have to succeed in order for the procedure to succeed) and from non-determinism (that only selects one branch).

The remaining operators are classical concurrency compositions. Rules Par (1)-(2) in Table~\ref{tab:CAop2} model the parallel
composition operator in
terms of {\em interleaving}, since only one basic $\omega$-action
is allowed for each transition (i.e., for each unit of time). This
means that the access to the shared AF $F$ is granted to one process at a time. However, time
passes for all the processes appearing in the $\|$ context
at the external level, as shown by rule Par (1), since
$\tau$-actions are allowed together with a $\omega$-action. On the
other hand, as shown by rule Par (2), a parallel component is allowed to proceed in
isolation if and only if the other parallel component cannot
perform a $\tau$-action.
To summarise, we adopt
maximum parallelism for time elapsing (i.e., $\tau$-actions) and an
interleaving model for basic computation steps (i.e., $\omega$-actions). 
By transition rules, an agent in a parallel composition obtained through $\|$ succeeds only if all the agents succeed. 
The parallel composition operator enables the specification of complex concurrent argumentation processes: for example, a debate involving many agents that asynchronously provide arguments can be modelled as a parallel composition of add operations performed on the knowledge base.
Any agent composed through $+$ (rules NDt (1)-(3)) is chosen if its guards succeed; the existential quantifier $\exists_x A$ behaves like agent $A$ where variables in $x$ are local to $A$.
Finally, in the procedure call (rule PrC), we consider the clause $p(x)::A$ present in the context $C$, in which $p(x)$ is executed. The parameter $x$ can be an argument, a label among \lin, \lout\ and \lundec, a semantics $\sigma$, or an instant of time. The procedure call can be extended if necessary to allow more than one parameter.
In rules HVa (Hidden Variables) and PrC, $A[x/y]$ denotes the agent obtained from $A$ by replacing variable x for variable y.

Using the transition system described in the rules of Tables~\ref{tab:CAop1} and \ref{tab:CAop2}, we can now define our notion of observables, which considers the traces based only on $\omega$-actions and whose components are AFs, of successful or failed terminating computations that an agent $A$ can perform for each {\itshape tcla} process $P=let~C~in~A$. 

\begin{definition}[Observables for {\itshape tcla}]
	Let $P=let~C~in~A$ be a {\itshape tcla} process. We define
	\[\begin{array}{lll}
		{\cal O}_{io}(P)&= \lambda F. & \{ F_1 \cdots F_n \cdot \mathit{ss} \mid
		F=F_1 \mbox{ and }\langle A, F_1\rangle \stackrel{\omega}{\longrightarrow} ^*\langle 
		success,  F_n \rangle\} \ \cup \\
		&  &  \{ F_1 \cdots F_n \cdot  \mathit{ff} \mid
		F=F_1 \mbox{ and }\langle A, F_1\rangle \stackrel{\omega}{\longrightarrow} ^*\langle 
		failure,  F_n \rangle\}. \\
	\end{array} 
	\]
\end{definition}

\section{Modelling Debates and Dialogue Games in {\itshape tcla}}\label{sec:applications}
This section provides two possible applications of {\itshape tcla}. In the first example, we model a persuasion dialogue between several debating counterparts, while the second example presents the formalisation of a dialogue game between two participants.
We propose operational procedures to translate both types of argumentative interaction into {\itshape tcla} processes.

\subsection{Modelling a Debate}\label{sec:example}
A possible use case for {\itshape tcla} can be identified in modelling information sharing for common
resource management.
This problem can be instantiated as done in other works like~\cite{DBLP:conf/atal/EmeleNP11,pagetinformation} as a debate in a multi-agent environment where argumentation techniques are exploited for arriving at desirable outcomes.
We start from the scenario proposed by~\cite{pagetinformation}, where three counterparts debate on the use of fertilisers for oyster production.

\begin{example}\label{ex:debate}
	We have three agents: Alice (a farmer), Bob (an oyster farmer) and Carol (a state
	representative). They are debating on the impact of the fertilisers on the oysters, as reported in the following:
	\begin{itemize}[leftmargin=2.4em]
		\item \textbf{Alice}: \textit{using a lot of fertiliser helps to have a big yield} (argument $a$);
		\item \textbf{Bob}: \textit{using a lot of fertiliser pollutes the lake and harms the oyster} (argument $b$);
		\item \textbf{Carol}: \textit{using a lot of fertiliser increases the risk of control} (argument $c$);
		\item \textbf{Carol}: \textit{using more fertiliser than the norm implies a fine} (argument $d$);
		\item \textbf{Alice}: \textit{there is no risk of being controlled because of lack of means} (argument $e$);
		\item \textbf{Carol}: \textit{an important polluting event can lead to harden the norms} (argument $f$);
		\item \textbf{Alice}: \textit{lake pollution is not linked to pesticides} (argument $g$).
	\end{itemize}
	A total of seven arguments are presented, upon which the AF of Figure~\ref{fig:ex1} is built.
	The specifications of this debate make it fall into the category of deliberation dialogues --~\cite{walton1995commitment}, where two or more participants, potentially with disagreeing positions, discuss in order to decide what course of action to take in a given situation.
	
	\begin{figure}[htb]
		\centering
		\includegraphics[width=0.55\linewidth]{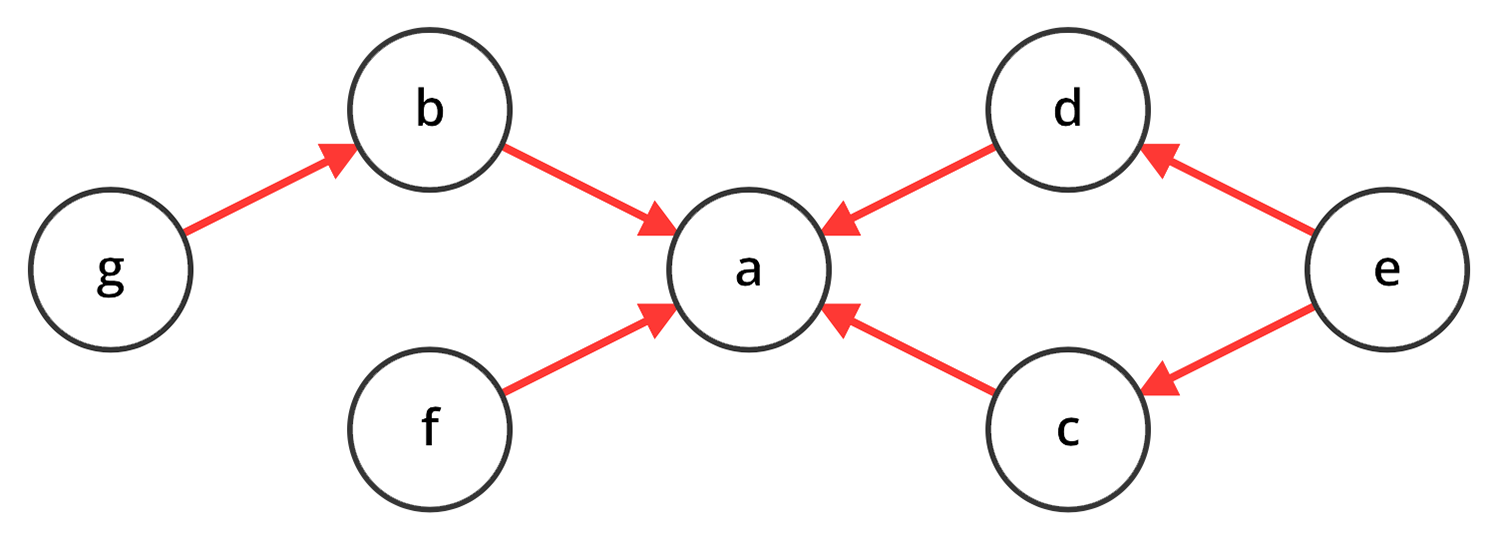}
		\caption{AF obtained starting from the arguments of Alice, Bob and Carol.}
		\label{fig:ex1}
	\end{figure}
	
\end{example}

Following \cite{DBLP:journals/logcom/Prakken05}, a finite debate $D$ between two or more participants (rational agents) with respect to an AF is a set of sequences of moves which respect a protocol and where several participants cannot speak at the same time.
All protocols are assumed to satisfy the following basic conditions: the arguments a participant could reasonably believe are non-conflicting, that is, a participant cannot reply to one’s own arguments; the arguments are not repeated, and the arguments can only attack other arguments that have already been introduced.
In order to formally model a finite debate, we introduce the definition of ordered sequence of arguments, where an argument can only attack arguments that precede it.

\begin{definition}
	Let 
	$F=\langle Arg, R\rangle$ be an AF and let  $D=\{s_1, \ldots, s_k\}$ be a set of possibly empty sequences of arguments, such that for $i\in [1,k]$,  $s_i=a_1^i \cdots a_{n_i}^i$. An ordered sequence associated to $D$ is a sequence of non repeated arguments
	$a_1 \cdots a_{n}$ such that 
	\begin{itemize}[leftmargin=2.4em]
		\item $\{a_1, \ldots, a_{n}\}=\bigcup_{i=1}^k (\cup_{j=1}^{n_i}\{a^i_j\})$,
		\item for each 
		$i \in [1,n]$ 
		$a_i^+ \subseteq \{a_1, \ldots, a_{i-1}\}$ and if $i>1$ then $a_i^+ \neq \emptyset$.
	\end{itemize}
\end{definition}

We denote by $ord(D)$ the set of the ordered sequences of $D$.
Now, we can formally define a debate, and in particular the three basic conditions associated with a protocol, as follows.
\begin{definition}[Debate]\label{def:debate}
	A debate $D$ in a multi-agent system, with respect to an AF
	$F=\langle Arg, R\rangle$, is a set of agents $D=\{Ag_1, \ldots, Ag_k\}$ where each agent $Ag_i$ is a sequence of arguments $s_i=a_1^i \cdots a_{n_i}^i$, such that 
	$\bigcup_{i=1}^k (\cup_{l=1}^{n_i}\{a^i_l\}) \subseteq Arg$ and the following holds
	\begin{itemize}[leftmargin=2.4em]
		\item for each $i \in [1,k]$, $l,r \in [1,n_i]$
		$(a_l^i, a_r^i) \not\in R$;
		\item for each $i,j \in [1,k]$, $l \in [1,n_i]$ and 
		$r \in [1,n_j]$
		if $i\neq j$ or $l\neq r$ then $a_l^i \neq a_r^i$;
		\item $ord(D) \neq \emptyset$.
	\end{itemize}
\end{definition}

\begin{definition}[Traces associated to a debate]\label{def:AFdebate}
	A trace of a debate $D$ is a sequence of pairs of arguments and attacks
	$\langle a_1, R_{|a_1}\rangle\cdots \langle a_n, R_{|a_n}\rangle$ such that $a_1 \cdots a_{n} \in ord(D)$. $tr(D)$ denotes the set of traces of the debate $D$.
	Moreover, we denote by 
	$TrF(D)$ the set of sequences of abstract frameworks associated
	to traces in $tr(D)$. Namely
	\[\begin{array}{lll}
		TrF(D)&=&\{\langle A_1, R_{1}\rangle\cdots \langle A_n, R_{n}\rangle
		\mid  \\
		&& \hspace*{1cm} \mbox{ there exists }\langle a_1, R_{|a_1}\rangle\cdots \langle a_n, R_{|a_n}\rangle \in tr(D)    \\
		&& \hspace*{1cm} \mbox{ such that for } l \in [1,n], \ 
		A_l=\{a_1, \ldots, a_l\} \mbox{ and } R_l= \cup _{j=1}^l R_{|a_j}\}
	\end{array} 
	\]
\end{definition}

\begin{example}\label{ex:debate1}
	By using the previous definition, we can model the debate in Example 
	\ref{ex:debate} by using the following three agents: 
	\begin{itemize}[leftmargin=2.4em]
		\item \textbf{Alice} = $a\cdot e \cdot g$;
		\item \textbf{Bob} = $b$;
		\item \textbf{Carol} = $c\cdot d \cdot f$.
	\end{itemize}
	Note that $a \cdot b \cdot c \cdots g \in ord(
	\{ \textbf{Alice}, \textbf{Bob}, \textbf{Carol}\}).$
\end{example}

Given the AF $F=\langle Arg, R \rangle$ and $a \in Arg$, we define the procedure 
\[\begin{array}{ll}
	wait(a) ::- & (check_{1}(a^{+},\{\}) \rightarrow  
	add(a,R_{|a})) 
	\ +_P  \\
	& check_{1}(\{\},\{\}) \rightarrow wait(a)
\end{array}
\]
where we use $add(Arg, R)$  as shortcut for $add(Arg, R)\rightarrow success$.
The procedure $wait(a)$ executes the check operation until $a^+$ (the set of arguments attacked by $a$) is contained in the set of arguments of the knowledge base, and then the agent $add(a,R_{|a} ) \rightarrow  success$  is executed. 
Practically, $wait(a)$ lets the action $add(a)$ wait for the presence of the arguments attacked by $a$ in the common store. 

In the following, $\varepsilon$ denotes an empty sequence
and given a possibly empty sequence $s= a_1 \cdots a_n$  of elements we denote by $rd(s)$ the sequence obtained from $s$ by removing duplicates of a same element.

\begin{definition}[Translation]
	The translation of a debate $D=\{Ag_1, \ldots, Ag_k\}$ in a multi-agent environment with respect to the AF $F=\langle Arg, R \rangle$, such that for each $i=1, \ldots, k$,  $Ag_i$ is the sequence of arguments $a_1^i \cdots a_{n_i}^i$ is defined as
	\[
	{\cal T}(D)={\cal T}(Ag_1, \ldots,Ag_k)=
	{\cal T}(a^1_1 \cdots a^1_{n_1})
	\|\cdots \|{\cal T}(a^k_1 \cdots a^k_{n_k})
	\]
	where
	\[
	{\cal T}(s) = \begin{cases}
		success &\  \text{if $s=\varepsilon$ }\\
		wait(a_1) \, \| \,
		{\cal T}(a_2 \cdots a_n) &\  \text{if $s=a_1 \cdots a_n$}
	\end{cases}
	\] 
\end{definition}

We can write a {\itshape tcla} program emulating such an exchange of arguments, using three agents in parallel to model the behaviour of Alice, Bob and Carol, respectively.
Each agent inserts the arguments at its disposal into the knowledge base through add operations. The first argument to appear in the debate is $a$, and since it does not attack any other argument, it can be directly added to the AF.
The arguments that come after and attack $a$, namely $b$, $c$, $d$ and $f$, are not brought forward before $a$ itself has been added. Indeed, although {\itshape tcla} allows one to add arguments and attacks to the knowledge base at separate times, in a debate we want arguments that come after $a$, namely $b$, $c$, $d$ and $f$, to be added together with their attacks toward $a$. Also the order in which the arguments are added must be respectful of the timing with which the debate between the three contenders takes place.
To ensure that those arguments will always be added after $a$, agents acting in place of Bob and Carol have to perform, beforehand, a check operation to verify whether $a$ belongs to the shared memory. Only once the check succeeds the agents can go on with the execution.
Analogously, Alice will check the arguments $c$ and $d$ before adding $e$, which attacks them, and the argument $b$ before adding $g$.
By using the previous definition, we can model the debate in Example 
\ref{ex:debate1} by the process
$\textbf{Alice} \,
\| \, \textbf{Bob}  \, \| \, \textbf{Carol}$, where 
\begin{itemize}
	\item \textbf{Alice} = $wait(a)\, \| \, wait(e) \, \| \, wait (g)$;
	\item \textbf{Bob} = $wait(b)$;
	\item \textbf{Carol} = $wait(c)\, \| \, wait(d)\, \| \, wait(f)$.
\end{itemize}

The following theorem, whose proof is immediate by definition of translation, states the translation of a debate $D$
produces a {\itshape tcla} process $P$ such that the set of traces of the debate $D$ is the set of successful sequences of the computed AF by $P$, without duplicates, starting from 
the store $F_0= \langle \emptyset, \emptyset \rangle$.

\begin{theorem}
	The translation ${\cal T}$ of a
	debate $D=\{Ag_1, \ldots, Ag_k\}$ in a multi-agent environment, with respect to the AF $\langle Arg, R\rangle$ 
	produces a {\itshape tcla} process $P$ such that
	\[TrF(D) = \{rd(F_1 \cdots F_n) \mid  \langle \emptyset, \emptyset \rangle \cdot F_1 \cdots F_n \cdot \mathit{ss} \in {\cal O}_{io}(P)\}
	\]
\end{theorem}

Note that we can give alternative translations of the same debate, as shown by the {\itshape tcla} process in Table~\ref{tab:ex1}. Since parallel executions are handled via interleaving, only one agent will operate on the knowledge base at a time, simulating the alternation of the three counterparts in exchanging arguments during the debate.
Check operations, in particular, allow agents to wait for their turn to ``speak''. In this example, we specify a timeout of $9$ instants of time, meaning that the check will be repeated up to $9$ times until it is either satisfied or expired.
In our case, checks can always be successful before their timeouts. Note that a shorter timeout cannot guarantee the successful termination of all check operations.
Different solutions can also be implemented. For instance, arguments $c$ and $d$ could be added with two distinct operations or together with $f$.

\begin{table}[htb]
	\centering
	\begin{align*}
		&add(\{a\},\{\}) \rightarrow 
		\\ 
		& \hspace{1cm} check_9(\{c,d\},\{\}) \rightarrow 
		add(\{e\},\{(e,c),(e,d)\}) \rightarrow success \,\,\, \|_G\\
		&\hspace{1cm} check_9(\{b\},\{\}) \rightarrow
		add(\{g\},\{(g,b)\}) \rightarrow success  \, 
		\\
		\| \\
		&check_9(\{a\},\{\}) \rightarrow add(\{b\},\{(b,a)\}) \rightarrow 
		success \,
		\\ \| 
		\\
		&check_9(\{a\},\{\}) \rightarrow add(\{c,d\},\{(c,a),(d,a)\}) \rightarrow success \,\,\, \|_G \\
		&check_9(\{a\},\{\}) \rightarrow add(\{f\},\{(f,a)\}) \rightarrow success
	\end{align*}
	\caption{\textit{tcla} program realising the AF of Figure~\ref{fig:ex1}.}
	\label{tab:ex1}
\end{table}

\subsection{Modelling Dialogue Games}\label{sec:example1}

Dialogues can also be constrained to form dialogue games --~\cite{DBLP:books/sp/09/McBurneyP09}, adding rules that govern their unfolding.
In this section, we provide an example of how {\itshape tcla} programs can be used to model dialogue games involving two agents taking ``turns'' to assert their beliefs.
Different dialogue games have been developed in the area of computational dialectics.
We adopt a simple, popular game often used in the literature --~\cite{procaccia2005extensive,DBLP:books/daglib/0011565,yuan2007computer} -- in which the dialectical framework is defined as dialogues between
two players $\textbf{P}$ (the ``proponent'') and $\textbf{O}$ (the ``opponent''), each of which is referred to as the other’s ``counterpart''.
We use a scenario given by~\cite{DBLP:journals/ijar/PolbergH18}, where two individuals $\textbf{P}$ and $\textbf{O}$ exchange arguments $a, b, \ldots$ about the safety of the children flu vaccine.

\begin{example}\label{ex:game}
	Let us consider the following dialogue game between two agents. The dialogue starts with $\textbf{P}$ (the proponent) claiming that the vaccine is not safe, to which $\textbf{O}$ (the opponent) objects and the discussion proceeds to revolve around whether it contains mercury-based compounds or not. 
	
	\begin{itemize}[leftmargin=2.4em]
		\item \textbf{P}: \textit{The flu vaccine is not safe to use by children} (argument $a$);
		\item \textbf{O}: \textit{The flu vaccine contains no poisonous components and is safe} (argument $b$);
		\item \textbf{P}: \textit{The vaccine contains some mercury compounds.
			The mercury compounds are poisonous, and therefore the vaccine is not safe to use} (argument $c$);
		\item \textbf{O}: \textit{The child vaccine does not contain any mercury compounds. The virus is only accompanied by stabilisers and possibly trace amounts of antibiotics used in its production} (argument $d$);
		\item \textbf{P}: \textit{The vaccine contains a preservative called thimerosal which is a mercury-based compound} (argument $e$);
		\item \textbf{O}: \textit{Children receive the nasal spray vaccine, and thimerosal has been removed from it over 15 years ago} (argument $f$).
	\end{itemize}
	
	According to~\cite{walton1995commitment}, the proposed example can be seen as a persuasion dialogue game since each participant tries to convince the other that his/her own position on the vaccine is correct.
\end{example}

A game begins with the player $\textbf{P}$ moving an initial
argument $x$ to be tested. The opponent $\textbf{O}$ and 
$\textbf{P}$ then take turns in moving
arguments that attack their counterpart’s last move.
The dialogue game is formalised as follows.

\begin{definition}
	Let $F=\langle Arg, R\rangle$ be an AF. A dialogue game is a quadruple
	$G= \langle F, d, Rl, Pls\rangle$,
	where $d$ (the dialogue history) is a possibly infinite sequence of moves  $m_0,m_1, \ldots$
	which contains the moves made by game participants,
	$Rl$ is a set of rules regulating players to make a move, and $Pls$ is
	a pair of players $\{P_0, P_1\}$, with $P_0$ and $P_1$ representing the proponent $\textbf{P}$ and the opponent $\textbf{O}$, respectively.
	The dialogue $d$ is such that:
	
	\begin{itemize}[leftmargin=2.4em]
		\item $\varepsilon$ denotes the empty dialogue and, if $d$ is not empty, $m_0$ the ``initial move'';
		\item  each $m_i$ is of the form $x_{Pl}$ where $x \in Arg$ is the argument moved in $m_i$, denoted by $arg(m_i)$, and 
		$Pl \in \{\textbf{P}, \textbf{O}\}$ is the player of $m_i$, denoted $pl(m_i)$.
	\end{itemize}
	
	In the following, given $d=m_0,m_1, \ldots$ we denote by 
	$arg(d)$ the possibly infinite set of arguments $\cup_{i\geq 0\ } arg(m_i)$.
	Finally the set $Rl$ contains six rules:
	
	\begin{enumerate}[leftmargin=2.4em]
		\item  First move in $d$ is made by $\textbf{P}$: $pl(m_0) = \textbf{P}$;
		\item Players take turns making moves: $pl(m_i) \neq pl(m_{i+1})$;
		\item Players cannot repeat a move, namely the same argument cannot appear twice in a game: if $i\neq j$ and $m_i, m_j \in d$ then $arg(m_i)\neq arg (m_j)$;
		\item Each move has to attack (defeat) only the previous (target) move: for each $i \geq 1$, $(arg(m_i), arg(m_{i-1}))\in R$, and there is no $j \in [0, i-2]$ such that $(arg(m_i), arg(m_{j}))\in R$;
		\item The game is ended if no further moves are possible: 
		$d=m_0,m_1, \ldots, m_n$ and there is no $a \in Arg \setminus arg(d)$ such that $(a, arg(m_n)) \in R$;
		\item The winner of an ended game is the player that makes the final
		move. In this case, the finite dialogue  $d =m_0,m_1, \ldots, m_n$ is said to be won by $P_{winner}=pl(m_n) =P_{n \,mod\, 2}$ and we denote by $result(d)$ the AF 
		$\langle arg (d), \{(arg(m_i), arg (m_{i-1}))\mid i \in [1,n]\}.$
	\end{enumerate}
	
\end{definition}

\begin{example}\label{ex:game1}
	By using the previous definition, we can model the dialogue in Example \ref{ex:game} by 
	$G= \langle F, d, Rl, Pls\rangle$ with
	$F=\langle Arg, R\rangle$, where 
	\[ \begin{array}{lll}
		Arg & = & \{a,b,c,d,e,f\},  \\
		R & =  & \{(a,b), (b,a), (b,c), (c,b), (c,d), (d,c), (d,e), (e,d), (e,f), (f,e)\}.
	\end{array}
	\]
	and 
	$d=a\cdot b\cdot c \cdots f$.
\end{example}

We can write a {\itshape tcla} program emulating a game in the AF $\langle Arg, R\rangle$, using two agents in parallel to model the behaviour of \textbf{P} and \textbf{O}, respectively.
Each agent inserts the arguments at its disposal into the knowledge base through an add operation. 
Therefore, if the argument of the move $m_i$ of the agent $Pl$ attacks the argument in the move $m_{i-1}$ 
then $Pl$ inserts in the knowledge base the argument $arg(m_i)$ and the attack $(arg(m_i), arg(m_{i-1}))$.
In the following definition, as in the previous section, we use $add(Arg, R)$  as a shortcut for $add(Arg, R)\rightarrow success$. Moreover, given a dialogue $d=m_0 \cdot m_1\cdot m_2 \cdots$, we denote by $even(d)$ ($odd(d)$) the sequence of arguments $arg(m_0) \cdot arg(m_2)\cdots$ $(arg(m_1) \cdot arg( m_3)\cdots).$

\begin{definition}[Translation]\label{def:translationGame}
	Let $turn_i, finish_i\not \in Arg$ for $i=0,1$ and let $hold(a)$ a shorthand for 
	$check_4(\{a\},\{\})\rightarrow rmv(\{a\},\{\})$.
	The translation of a finite dialogue $d$ in the dialogue game
	$G= \langle F, d, R, Pls\rangle$, is defined as follows.
	\[
	{\cal F}(d)=\begin{cases}
		\textbf{success} &\  \text{if $d=\varepsilon$ }\\
		{\cal F}_ \textbf{P}(even(s))
		\|{\cal F}_ \textbf{O}(odd(s))&\  \text{otherwise}
	\end{cases}
	\]
	where 
	\[{\cal F}_ \textbf{P}(a \cdot s')=
	add(\{a, turn_1\},\{\}) \rightarrow  {\cal F}'_0(s');
	\]
	\[{\cal F}_  \textbf{O}(s)=
	\begin{cases}
		hold(turn_1)\rightarrow add(\{finish_1\},\{\}) &\  \text{if $s=\varepsilon$ }\\
		hold(turn1)\rightarrow add(\{a, turn_0\},\{R_{|a}\})  \rightarrow {\cal F}'_1(s')&\  \text{if $s=a\cdot s'$}
	\end{cases}
	\]
	and for $i=0,1$,    
	\[
	{\cal F'}_i(s)=\begin{cases}
		hold(finish_{i+1 mod 2})\rightarrow \textbf{success}
		\ +_p \  hold(turn_{i})\rightarrow add(\{finish_i\},\{\}) \\  
		\hspace{8cm} \text{if $s=\varepsilon$ }\\
		hold(turn_{i})
		\rightarrow add(\{a, turn_{i+1 mod 2}\},R_{|a}) 
		\rightarrow {\cal F'}_i(s') \\  
		\hspace{8cm}  \text{if $s=a \cdot s'$} 
	\end{cases}
	\] 
\end{definition}

The following theorem, whose proof is immediate by definition of translation, states the translation of a dialogue $d$
produces a {\itshape tcla} process $P$ such that for each finite successful finite trace computed by $P$, starting from 
the empty store $F_0= \langle \emptyset, \emptyset \rangle$, 
$result(d)$ is the last computed AF of the trace itself.

\begin{theorem}
	The translation ${\cal F}$ of a finite dialogue game $G= \langle F, d, R, Pls\rangle$  produces a {\itshape tcla} process $P$ such that for each sequence 
	$F_0 \cdot F_1 \cdots F_m \cdot \mathit{ss} \in {\cal O}_{io}(P)$,
	where $F_0=\langle \emptyset, \emptyset \rangle$
	\begin{itemize}[leftmargin=2.4em]
		\item $result(d)=F_m$ and
		\item for $i=0,1$, $P_{winner}=P_i$ if and only if there exists 
		$j \in [0,n]$ such that $F_j=\langle A_j,R_j \rangle$ and 
		$finish_{(i+1) \, mod \, 2} \in A_j$.
	\end{itemize}	
\end{theorem}

Note that the previous translation can be extended in a straightforward  way in order to also model:
\begin{itemize}
	\item dialogue games with more than two players,
	\item game trees --~\cite{procaccia2005extensive,rahwan2009argumentation} -- representing sequences of actions agents may take, and
	\item turn-taking functions that determine how the turn shifts starting from the current player and from the abstract framework produced by the portion of the dialogue that has already taken place.
\end{itemize}

\begin{example}
	By using Definition \ref{def:translationGame} we can model the dialogue $d$ in the Example \ref{ex:game} by the process: 
	\[{\cal T}(d)=
	{\cal T}_ \textbf{P}(a\cdot c\cdot e)
	\|{\cal T}_ \textbf{O}(b\cdot d \cdot f)
	\]
	where
	\[\begin{array}{ll}
		{\cal T}_ \textbf{P}(a\cdot c\cdot e)\ = & add(\{a, turn_1\},\{\}) \rightarrow  
		\\
		& hold(turn_{0})
		\rightarrow  add(\{c, turn_{1}\},\{(c,b)\})  \rightarrow\\
		& hold(turn_{0})
		\rightarrow add(\{e, turn_{1}\},\{(e,d)\})
		\rightarrow \\
		& ( hold(finish_{1})\rightarrow \textbf{success}
		\ +_p \  hold(turn_{0})\rightarrow add(\{finish_0\}))
	\end{array}
	\]
	and 
	\[\begin{array}{ll}
		{\cal T}_ \textbf{O}(b\cdot d\cdot f)\ = & hold(turn1)\rightarrow add(\{b, turn_0\},\{(b,a)\})  \rightarrow 
		\\
		& hold(turn1)\rightarrow add(\{d, turn_0\},\{(d,c)\})  \rightarrow 
		\\
		& hold(turn_{1})
		\rightarrow  add(\{f, turn_{0}\},\{(f,e)\})  \rightarrow\\
		& ( hold(finish_{0})\rightarrow \textbf{success}
		\ +_p \  hold(turn_{1})\rightarrow add(\{finish_1\}))
	\end{array}
	\]
\end{example}


\section{{\itshape tcla} Simulator}\label{sec:implementation}

We developed a working implementation for the interleaving version of {\itshape tcla}.
Some of the operations had their syntax translated (see Table~\ref{tab:translation}) to enable users to specify {\itshape tcla} programs manually.
In this section, we describe the details of our implementation, also comparing the interleaving version of {\itshape tcla} with the maximum parallelism version.

\begin{table}[htb]
\centering
\caption{Implementation of {\itshape tcla} operations}
\label{tab:translation}
\begin{tabular}{rl}
	\hline\hline
	{\itshape tcla} syntax & Implementation\\
	\hline
	$add(Arg,R)$ & \verb#add(Arg,R)#\\
	$rmv(Arg,R)$ & \verb#rmv(Arg,R)#\\
	$check_t(Arg,R)$ & \verb#check(t,Arg,R)#\\
	$\mbox{c-}test_t(a,l,\sigma)$ & \verb#ctest(t,{a},l,#$\sigma$\verb#)#\\
	$\mbox{s-}test_t(a,l,\sigma)$ & \verb#stest(t,{a},l,#$\sigma$\verb#)#\\
	$E + \dots + E$ & \verb#sum(E,...,E)#\\
	$E \|_G \dots \|_G E$ & \verb#gpar(E,...,E)#\\
	$E+_P E$ & \verb#(E)+P(E)#\\
	\hline\hline
\end{tabular}
\end{table}

The simulator's core consists of a Python script that covers three fundamental tasks: it serves as an interpreter for the {\itshape tcla} syntax, executes programs taken in input, and communicates with a web interface.
The interpreter is built using ANTLR,\footnote{ANTLR website: \url{https://www.antlr.org}.} a parser generator for reading, processing, executing, and translating structured text.
We start with a grammar file defining the constructs in Table~\ref{tab:CA}. Any source program, then, is parsed according to the grammar, and a parse tree is generated.
ANTLR provides two ways to traverse the parse tree: via a listener or a visitor.
The most significant difference between these two mechanisms is that listener methods are called independently by a built-in ANTLR object, while visitor methods must traverse the tree by recursively visiting the children of visited nodes with explicit visit calls.
Since we want to implement guards in our language, we must have the possibility of deciding which part of the tree will be visited, making our choice fall on the visitor approach.

Each node of the parse tree corresponds to one operation to perform, whose behaviour is defined in a dedicated Python class. Hence, visiting the parse tree is equivalent to executing the corresponding program.
The root of the parse tree is always a \textit{visitPrg} node, which calls the visit on its children, collects the results and returns the final output.
Below, we provide details on the visiting functions for the various node types.

The terminal nodes are \textit{visitSuc} and \textit{visitFlr}, which represent the leaves of the tree and correspond to \verb#success# and \verb#failure# agents, respectively.

\textit{visitAdd} and \textit{visitRmv} nodes implement \verb#add(Arg,R)# and \verb#rmv(Arg,R)# operations, respectively. They modify the knowledge base by adding/removing part of the framework, always succeeding and continuing on their children. For adding an attack $(a,b)$, arguments $a$ and $b$ must be contained in the shared memory when \textit{visitAdd} is performed. The two arguments can be introduced in the same step in which the attack between them is added. \textit{visitRmv}, then, also succeeds if the specified arguments and attacks are not in the AF; in that case, the AF is left unchanged.

\textit{visitChk} checks if a given set of arguments and attacks belongs to the knowledge base at time $t$ as per the \verb#check(t,Arg,R)# operator. In case of success, the visit proceeds to the consequent action. On the other hand, when the knowledge base does not contain the specified parts of AF, the timeout is decreased and the check repeated. When the timeout reaches zero, \textit{visitChk} fails.

\textit{visitTcr} (\verb#ctest(t,{a},l,#$\sigma$\verb#)#) and \textit{visitTsk} (\verb#stest(t,{a},l,#$\sigma$\verb#)#) call the ConArg\footnote{ConArg is a Constraint Programming-based tool for solving computational argumentation problems. It is available online at the following webpage: \url{https://conarg.dmi.unipg.it/}.} solver to credulously/sceptically test the acceptability of a given argument $a$, with respect to a semantics $\sigma$ at a time $t$. The functions repeat the verification until either the test succeeds or the timeout reaches zero. In the latter case, both constructs return \verb#failure#. 

A \textit{visitNdt} node implements \verb#sum(E,...,E)#, which is a concatenation of $+$ operators, inspecting the guards of all its children expressions and randomly selecting one branch to execute among the possible ones. If no guards are found with satisfiable conditions, all timeouts are decreased, and the conditions are rechecked in the next step. Expressions with expired timeouts are discarded, and if no expression can be executed before the last timeout expires, the construct fails.

\textit{visitIte} behaves like the if-then-else construct \verb#(E)+P(E)#. The expressions are handled in the same order in which they are specified. If the first expression succeeds, \textit{visitIte} succeeds without executing the second one. If the first expression fails, the second one is executed. If also the second expression fails, the construct fails; otherwise, it succeeds.

The node \textit{visitPar} starts separated threads to execute two parallel agents composed through the operator $\parallel$. It returns true if both agents succeed or false as soon as one action fails.

\textit{visitGpa} implements \verb#gpar(E,...,E)#, namely a concatenation of $\|_G$ operators. It cycles through its children, starting a new thread for every expression found with a satisfiable guard. If all the executed expressions succeed, the construct succeeds. In case no expression can be executed, all timeouts are decreased, and expressions with expired timeouts are discarded. The process is repeated until either all the expressions have been executed or discarded for timeout or when one of the expressions terminates with a failure.

The parallel execution of $\omega$-actions is handled through interleaving: only one
$\omega$-action, i.e., one operation of the types \textit{visitAdd}, \textit{visitRmv}, \textit{visitChk}, \textit{visitTcr} and \textit{visitTsk}
can be executed at each step. Such a behaviour is accomplished by means of a Python lock object which acts as a synchronisation primitive, entrusting the control of the shared memory to one action at a time. In detail, when an $\omega$-action is ready to be executed, it tries to acquire the lock object. If the object is unlocked, it immediately changes its status to locked, and the action continues its execution. Before proceeding to the subsequent step, the action releases the lock, which becomes unlocked again.
If, on the other hand, the object is locked, the action cannot be executed
(because another $\omega$-action has already been granted such privilege upon the acquisition of the lock) and, thus, it will be postponed to the next step.
Practically, we rewrite the parse tree of the program so that each node representing an $\omega$-action $A$ that cannot be executed at a given step $s$ is assigned a child node which is a clone of $A$ itself, except for possible timeouts (only present in \textit{visitChk}, \textit{visitTcr} and \textit{visitTsk} operations), that are decreased by one.
Failed attempts of execution also consume a unit of time: when the condition of a guarded $\omega$-action is not satisfied, its timeout is decreased, and the execution is postponed by one step.

Differently from $\omega$-actions, $\tau$-actions do not directly interact with the underlying knowledge base, as they are used to make time pass for timeout operations.
Several $\tau$-actions can be executed concurrently with an $\omega$-action at each step.
To obtain maximum parallelism for $\tau$-actions, we synchronise the threads that implement the agents by keeping track of CPU time elapsing in each parallel branch of the execution. We have equipped the interpreter with an internal scheduler that manages the execution of the various actions. In each step, the scheduler executes $\tau$-actions of parallel threads together with a single $\omega$-action. In this way, only one agent at a time has access to the shared memory, while timeouts in all parallel branches are decremented at the same instant t.
Since the execution of all operations is governed by a scheduler function that synchronises the threads, the $\tau$-actions in a certain step will be executed in parallel within the interpreter, which will see each of these actions terminated at the same instant $t$. However, the $\tau$-actions will not be executed with true parallelism by the processor, which can only perform one operation at a time.

\begin{figure}[htb]
\centering
\includegraphics[width=\linewidth]{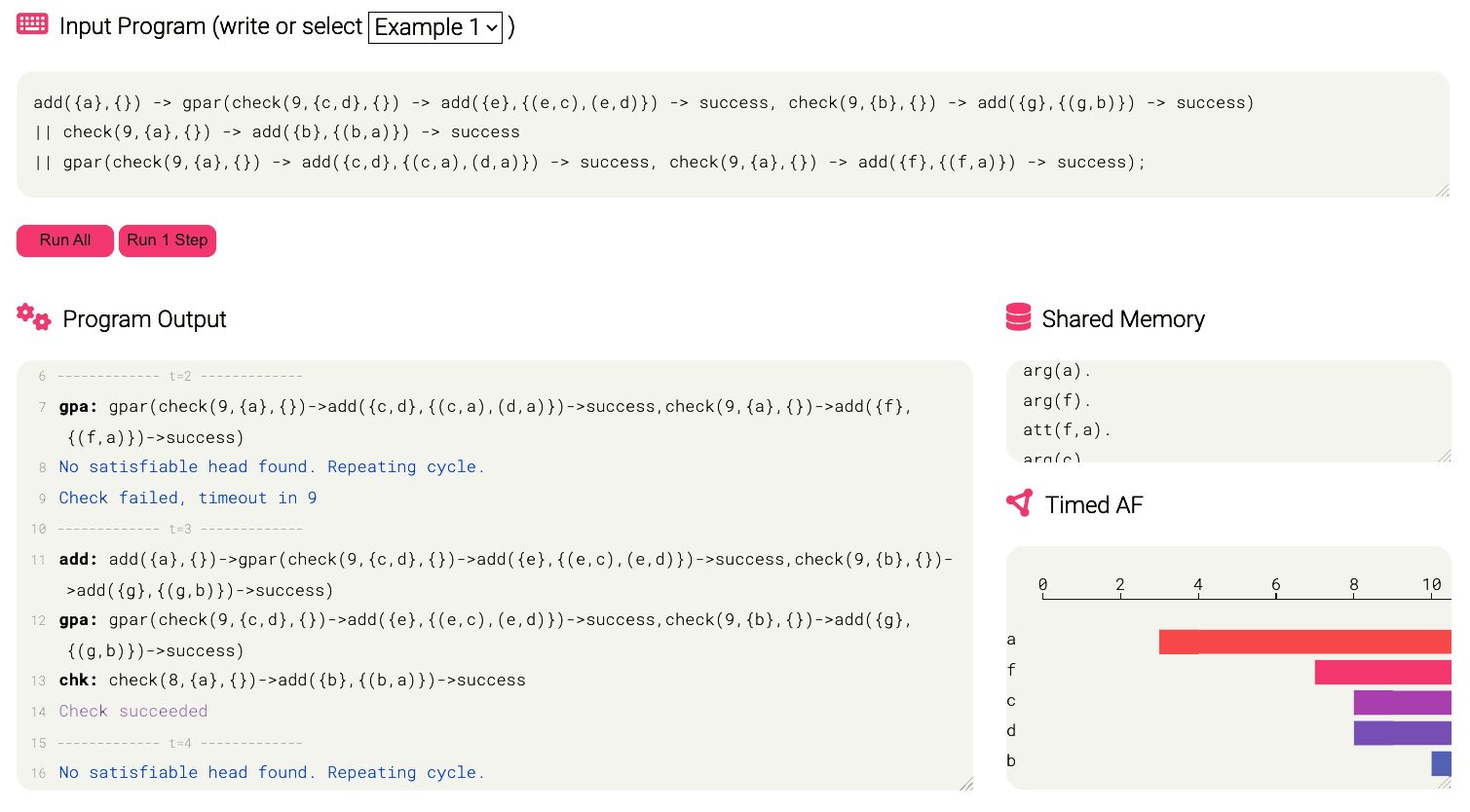}
\caption{Example of execution via the web interface of the {\itshape tcla} program given in Table~\ref{tab:ex1}.}
\label{fig:interface}
\end{figure}

The input program is provided to the Python script through a web interface\footnote{The web interface can be tested at the following link: \url{https://conarg.dmi.unipg.it/tcla-i}.} (see Figure~\ref{fig:interface}), developed in HTML and JavaScript.
After the program has been executed, its output is also shown within the interface.
We have two main areas, one for the input and the other for the output. First, the user enters a program in the dedicated text box (either manually or by selecting one of the provided examples), after which there are two ways to proceed. By clicking on the ``Run All'' button, the whole program is executed at once, and the final result is displayed in the output area. Alternatively, by clicking on the ``Run 1 Step'' button, it is possible to monitor the execution step by step.


The interface communicates with the underlying Python engine through an Ajax call which passes the program as a parameter and asynchronously retrieves the output.
After the execution of (a step of) the program, three different components are simultaneously visualised in the output area: the program output, the state of the shared memory and a timeline representing the behaviour of arguments during the time.
The program output box shows the results of the execution, divided by steps. The beginning of each step is marked by a separating line explicitly showing the step number.
The shared memory box is updated after each step of the execution and shows the AF used as the knowledge base.
Finally, the bottom-left box contains the visual representation of arguments during the time and shows the temporal evolution of the AF used by the {\itshape tcla} program. Time is reported on the $x$ axis, and each bar of the timeline shows the intervals of time during which an argument is contained in the shared memory.

We now compare the interleaving-based implementation of {\itshape tcla} reported in this paper with the approach presented in~\cite{tcla1,app2} where maximum parallelism is used for allowing parallel execution of an infinite number of concurrent agents.
We use an example to illustrate how the two versions handle actions in parallel constructs differently.

\begin{example}\label{ex:comp}

Consider the following {\itshape tcla} program, which involves three parallel agents. Each of those agents performs an addition operation of a different argument and then succeeds.

\begin{verbatim}
	add({a},{}) -> success || 
	add({b},{}) -> success || 
	add({c},{}) -> success;
\end{verbatim}

Since the two implementations share the same syntax for the various language constructs, the interpreter will produce the same parse tree for both the interleaving and the maximum parallelism approach.
The parse tree generated for the program used in this example is depicted in Figure~\ref{fig:parse_tree}.

\begin{figure}[htb]
	\centering
	\includegraphics[width=\linewidth]{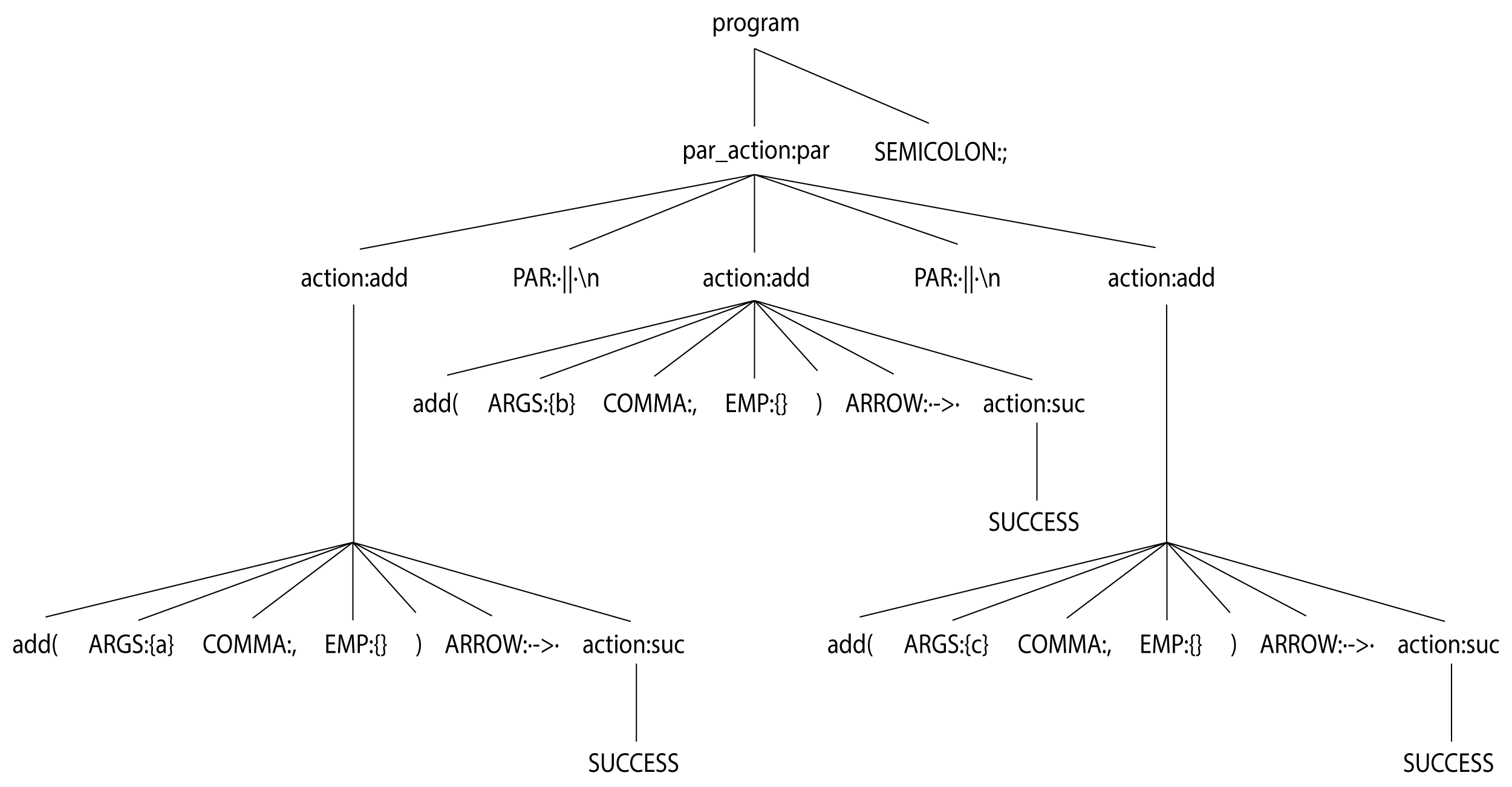}
	\caption{Parse tree for the {\itshape tcla} program of Example~\ref{ex:comp}.}
	\label{fig:parse_tree}
\end{figure}

We distinguish three main parallel branches in this tree, each containing one addition operation for argument $a$, $b$ and $c$, respectively.
When executed with the maximum parallelism version of the simulator, the three additions are performed at the same instant of time $t=0$, as shown in Figure~\ref{fig:taf_comp} right, and arguments $a$, $b$ and $c$ will be all available in the shared memory starting from $t=1$.
On the other hand, when interleaving is used, only one addition operation is executed at each instant, so the three arguments are inserted in the shared memory at different times. In this example, \verb#add({a},{})# is executed at $t=0$, \verb#add({b},{})# at $t=1$, and \verb#add({c},{})# at $t=2$.

\begin{figure}[htb]
	\centering
	\includegraphics[width=\linewidth]{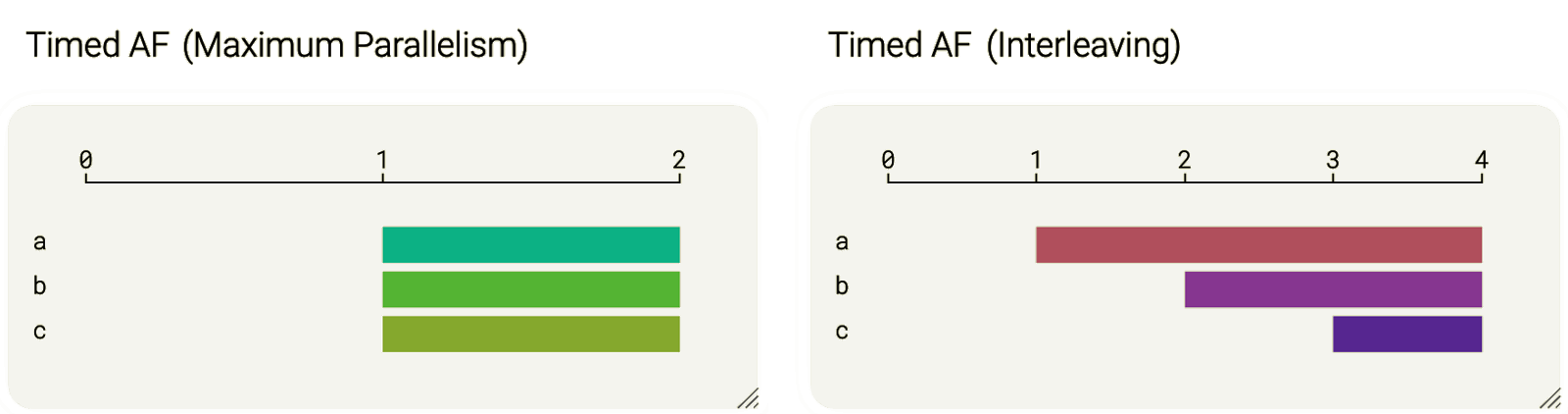}
	\caption{Availability of arguments in shared memory for the {\itshape tcla} program of Example~\ref{ex:comp} according to execution with maximum parallelism (left) and interleaving (right).}
	\label{fig:taf_comp}
\end{figure}

\end{example}

The two versions of {\itshape tcla} illustrated in Example~\ref{ex:comp} are designed to meet different needs in modelling interactions between argumentative agents. Therefore, they both have benefits and disadvantages depending on the use case.
When maximum parallelism is employed, different agents may perform several concurrent actions simultaneously. This can be particularly useful in multi-agent systems where many components compete for the allocation of resources through, for example, negotiation protocols. In this case, running multiple parallel processes offers two advantages: on the one hand, it increases the system's efficiency, and on the other hand, it enables the simulation of concurrent interactions between intelligent agents.
However, this approach is less suitable for modelling typical human interactions, such as debate, in which the interaction between the various agents is assumed to take place in a controlled manner. For example, in the dialogue games introduced in the previous section, the participants act according to a turn-based model. For this type of application, which is closer to the human way of reasoning, the interleaving version of {\itshape tcla} provides a better executing tool than the version with maximum parallelism.

\section{Related Work}\label{sec:related}

Time is an important aspect to consider when representing processes that take place in the real-life, and much effort has been put into studying models that incorporate the temporal dimension.
In this section, we report on some significant works in the literature that deal with both the dynamics of argumentation and, more broadly, languages and formalisms that use the notion of time.


A formalism for expressing dynamics in AFs is defined by~\cite{rotstein_abstract_2008} as a \textit{Dynamic Argumentation Framework} (DAF).
This framework allows for instantiating Dung-style AFs by considering ``evidence'' (a set of arguments to adopt) from a universe of arguments.
DAF generalises AFs by adding the possibility of modelling changes but, contrary to our study, it does not consider how such modifications affect the semantics and does not allow to model the behaviour of concurrent agents.

In our model, AFs are equipped with a universe of arguments that agents use to insert new information into the knowledge base. The problem of combining AFs, i.e., merging arguments and attacks of two different AFs, is addressed in a paper by~\cite{DBLP:journals/ai/BaumeisterNRS18}, that studies the computational complexity of verifying if a subset of arguments is an extension for a certain semantics in Incomplete Argumentation Frameworks. The incompleteness is considered both for arguments and attacks. Similarly to our approach, arguments and attacks can be brought forward by agents and used to build new acceptable extensions. On the other hand, the authors focus on a complexity analysis and do not provide implementations for the merging.

In the context of Argumentation Theory, it is reasonable to assume that the interaction between entities is regulated by the passing of time --~\cite{DBLP:conf/comma/MannH08,DBLP:conf/argmas/MarcosFS10}.
Timed Abstract Argumentation Frameworks (TAFs) --~\cite{DBLP:journals/isci/BudanLCS15,DBLP:conf/ecai/CoboMS10} -- have been proposed to meet the need for including the notion of time into argumentation processes.
The existence of arguments in a TAF is regulated by a function that determines the exact intervals of time in which every argument is available within the framework. In a previous work --~\cite{tcla1}, we used {\itshape tcla} constructs to dynamically instantiate a TAF. The interleaving approach we propose with the current work, however, is not suitable for that task: since only one agent can interact with the store at once, it is not possible to model a TAF in which, for instance, two different arguments are added and removed in the same instant of time.


The Linear Temporal Logic (LTL) was first proposed by~\cite{DBLP:conf/focs/Pnueli77} for the formal verification of concurrent systems. LTL is a modal logic that can express properties of programs over time through the use of two fundamental operators: \textit{next}, which says that a formula has to hold at the next state of the execution, and \textit{until}, expressing that a formula has to hold at least until another formula becomes true.
LTL differs from {\itshape tcla} both syntactically and semantically. In particular, contrary to {\itshape tcla}, LTL allows propositional variables and modal operators to express time-related properties.

Difference Logic can express properties of timed automata and is used to approach time-related problems like verification of timed systems and scheduling.
In the context of the Satisfiability Modulo Theories (SMT) problem --~\cite{DBLP:series/faia/BarrettSST21}, which concerns the satisfiability of formulas with respect to some background theory, a solver for Difference Logic can be obtained by using the DPLL(T) architecture proposed by~\cite{DBLP:journals/jacm/NieuwenhuisOT06}, where the T parameter is instantiated with a theory of integers that allow handling time.
However, SMT does not allow for the encoding of interactions between arguing agents, contrary to the system we propose, which instead has operators appropriately designed for this use.

To deal with quantitative temporal information,~\cite{DBLP:journals/cacm/Allen83} proposes an Interval Algebra where entities like actions or events can be placed in a timeline, with pairs of endpoints setting entities beginning and end. Then, various configurations between those entities can be encoded and studied through a set of binary temporal relations which express, for example, whether an event occurred before, after or overlapping another.
Arguments shared by {\itshape tcla} agents are bound by a time duration and can be regarded as entities in Allen's framework. However, although we can express the temporal availability of arguments by adding and removing them from the shared memory at set instants of time, {\itshape tcla} neither uses explicit time intervals nor the relative availability between arguments.

A collection of process calculi is presented by~\cite{DBLP:conf/amast/BusiGZ00} as a solution for the lack of formal definitions of languages like Linda, JavaSpaes and TSpaes. In this regard, an operational semantics is introduced to enable formal reasoning and allow the systematic comparison of primitives with respect to their expressiveness.
Although the authors consider the passing of time (represented as divided into discrete intervals), time does not elapse for timeout constructs. In our work, instead, also timeout processes can make time pass.

Other works related to ours extend CC with timed constructs, also based on the hypothesis of bounded asynchrony --~\cite{DBLP:conf/coordination/BistarelliGMS08,DBLP:journals/iandc/BoerGM00}. In these approaches, time elapsing is measured by means of a global clock and each time instant is marked through action prefixing. The resulting timed languages are able to describe the behaviour of intelligent agents interacting within a dynamic environment. Apart from the different scope of application (\cite{DBLP:conf/coordination/BistarelliGMS08} deal with constraint systems), the main difference with our work lies in the fact that maximum parallelism is assumed for concurrent actions instead of interleaving.

Interleaving is also used to model parallel composition of actions in the context of a temporal logic based on CC --~\cite{DBLP:journals/tocl/BoerGM04}. The paper's primary purpose is to devise a logic for reasoning about the correctness of timed concurrent constraint programs. Indeed, the authors focus on providing soundness and completeness of a related proof system rather than modelling complex reasoning processes in multi-agent systems. Consequently, a significant difference with our work is that information is monotonically accumulated in the shared memory, as per classical CC tell operation, and cannot be retracted by the agents.

\cite{DBLP:journals/jair/FoxL03} propose PDDL2.1, a planning language capable of modelling domains with temporal features. The language allows for defining actions endowed with activation conditions and effects. Through a system of temporal annotations, it is then possible to specify whether each condition and effect associated with an action must be valid at the beginning, at the end, or during the entire duration of that action. Like {\itshape tcla}, PDDL2.1 allows the management of concurrent processes. However, being designed as an action-oriented planning language, it does not support argumentative interactions, like debates ad dialogue games, in multi-agent systems.

\section{Conclusion}\label{sec:conclusions}

In this paper, we presented a formalisation of {\itshape tcla} based on two kinds of actions, $\tau$-actions and $\omega$-actions, which realise time elapsing and computation steps, respectively.
Parallel composition of $\tau$-actions is handled through maximum parallelism, while, for $\omega$-actions, we adopt an interleaving approach. Indeed, it seems more natural for timeout operators not to interrupt the elapsing of time once the evaluation of a timeout has started. Clearly, one could start the elapsing of time when the timeout process is scheduled rather than when it appears in the top-level current parallel context. This modification could easily be obtained by adding a syntactic construct to differentiate active timeouts from inactive ones and changing the transition system accordingly.
One could also easily modify the semantics (both operational and denotational) to
consider a more liberal assumption which allows multiple ask
actions in parallel.
Along with our language, we have shown two examples of possible applications where we consider debates and dialogue games, respectively. In particular, we have provided procedures for translating these two types of interaction into {\itshape tcla} programs, where the interleaving used to manage the parallelism of omega actions allows for modelling the behaviour of agents acting in turn.
Finally, we compared the interleaving-based implementation of {\itshape tcla} with the maximum parallelism implementation presented in previous work, highlighting the differences in running parallel agents and discussing the advantages offered in different use cases.

In future work, we first plan to use existential quantifiers to extend our language by allowing the agents to have local stores.
Then, to test the capabilities of our language and evaluate its expressiveness, we also plan to conduct a study on strategic argumentation --~\cite{DBLP:journals/flap/GovernatoriMO21,DBLP:journals/ki/Thimm14}: intelligent agents could interact through {\itshape tcla} constructs and modify their shared memory to reach the desired outcome in terms of accepted arguments.
In addition to classical AFs, we also would like to investigate other kinds of frameworks we might use as a knowledge base for our agents (e.g., Bipolar Argumentation Frameworks --~\cite{DBLP:conf/ecsqaru/CayrolL05a}, Abstract Dialectical Frameworks --~\cite{DBLP:conf/kr/BrewkaW10}, and Extended Argumentation Frameworks --~\cite{DBLP:journals/ai/Modgil09}).
Such refined AFs are endowed with more complex structures for the arguments and the relations between them, allowing one to model different nuances of reasoning processes.

\bibliographystyle{plain}
\bibliography{bibliography}

\end{document}